\documentclass[runningheads,a4paper]{llncs}
\usepackage{times}
\usepackage{amsmath}
\usepackage{amssymb}
\usepackage{cite}
\usepackage{graphicx}
\usepackage{subfigure}
\usepackage{array}
\usepackage{url}
\usepackage{algorithm}
\usepackage{algorithmic}

\newtheorem{myDef}{Definition}

\begin{document}


\title{NLPMM: a Next Location Predictor with Markov Modeling}
\author{Meng Chen\inst{1} \and Yang Liu\inst{1} \and Xiaohui Yu\inst{1,2,}\thanks{Corresponding author.}}
\authorrunning{M.Chen et al.}

\maketitle

\begin{abstract}

In this paper, we solve the problem of predicting the next locations of the moving objects with a historical dataset of trajectories. We present a Next Location Predictor with Markov Modeling (\textit{NLPMM}) which has the following advantages: (1) it considers both individual and collective movement patterns in making prediction, (2) it is effective even when the trajectory data is sparse, (3) it considers the time factor and builds models that are suited to different time periods. We have conducted extensive experiments in a real dataset, and the results demonstrate the superiority of \textit{NLPMM} over existing methods.
\end{abstract}

\section{Introduction}

The prevalence of positioning technology has made it possible to track the movements of people and other objects, giving rise to a variety of location-based applications. For example, GPS tracking using positioning devices installed on the vehicles is becoming a preferred method of taxi cab fleet management. In many social network applications (e.g., Foursquare), users are encouraged to share their locations with other users. Moreover, in an increasing number of cities, vehicles are photographed when they pass the surveillance cameras installed over highways and streets, and the vehicle passage records including the license plate numbers, the time, and the locations are transmitted to the data center for storage and further processing. 


In many of these location-based applications, it is highly desirable to be able to accurately predict a moving object's next location.  Consider the following example in location-based advertising. Lily has just shared her location with her friends on the social network website. If the area she will pass by is known in advance, it is possible to push plenty of information to her, such as the most popular restaurant and the products on sale in that area. As another example, if we could predict the next locations of vehicles on the road, then we will be able to forecast the traffic conditions and recommend more reasonable routes to drivers to avoid or alleviate traffic jams.

Several methods have been proposed to predict next locations, most of which fall into one of two categories: (1) methods that use only the historical trajectories of individual objects to discover individual movement patterns \cite{xue2009traffic,jeung2008hybrid}, and (2) methods that use the historical trajectories of all objects to identify collective movement patterns   \cite{monreale2009wherenext,morzy2007mining}. The majority of the existing methods train models based on frequent patterns and/or association rules to discover movement patterns for prediction.

However, there are a few major problems with the existing methods. First, those methods focus on either the individual patterns or the collective patterns, but very often the movements of objects reflect both individual and collective properties. Second, in some circumstances (e.g., social check-in, and vehicle surveillance), the data points are very sparse; the trajectories of some objects may consist of only one record. One cannot construct meaningful frequent patterns with these trajectories. Finally, the existing methods do not give proper consideration to the time factor. Different movement patterns exist in different time, for example, Bob is going to leave his house. If it is 8 a.m. on a weekday, he is most likely to go to work. But if it is 11:30 a.m., he is more likely to  go to a restaurant, and he may go shopping if it is 3 p.m on weekends. Failing to take time factor into account would result in higher error rates in predicting the next locations.

To address those problems, we propose a Next Location Predictor with Markov Modeling (\textit{NLPMM}) to predict the next locations of moving objects given past trajectory sequences. \textit{NLPMM} builds upon two models: the Global Markov Model (\textit{GMM}) and the Personal Markov Model (\textit{PMM}). \textit{GMM} utilizes all available trajectories to discover global behaviours of the moving objects based on the assumption that they often share similar movement patterns (e.g., people driving from A to B often take the same route). \textit{PMM}, on the other hand, focuses on modeling the individual patterns of each moving object using its own past trajectories. The two models are combined using linear regression to produce a more complete and accurate predictor.  

Another distinct feature of NLPMM lies in its treatment of the time factor. The movement patterns of objects vary from one time period to another (e.g., weekdays vs. weekends). Meanwhile, similarities also exist for different time periods (e.g., this Monday and next), and the movement patterns of moving objects tend to be cyclical. We thus propose to cluster the time periods based on the similarity in movement patterns and build a separate model for each cluster. 

The performance of \textit{NLPMM} is evaluated in a real dataset consisting of the vehicle passage records over a period of 31 days  (1/1/2013 - 1/31/2013) in a metropolitan area \footnote{The name of the city is withheld due to the anonymity rule.}. The experimental results confirm the superiority of the proposed methods over existing methods.

The contributions of this paper can be summarized as follows.
\begin{itemize}
\item We propose a Next Location Predictor with Markov Modeling to predict the next location a moving object will arrive at. To the best of our knowledge, \textit{NLPMM} is the first model that takes a holistic approach and considers both individual and collective movement patterns in making prediction. It is effective even when the trajectory data is sparse.
\item Based on the important observation that the movement patterns of moving objects often change over time, we propose methods that can capture the relationships between the movement patterns in different time periods, and use this knowledge to build more refined models that are better suited to different time periods.
\item We conduct extensive experiments using a real dataset and the results demonstrate the effectiveness of \textit{NLPMM}.
\end{itemize}
The remainder of this paper is organized as follows. Section 2 reviews related work. Section 3 gives the preliminaries of our work. Section 4 describes our approach of Markov modeling. Section 5 presents methods that take the time factor into consideration. The experimental results and performance analysis are presented in Section 6. Section 7 concludes this paper.

\section{Related Work}

There have appeared a considerable body of work on knowledge discovery from trajectories, where a trajectory is defined as a sequence of locations ordered by time-stamps. In what follows, we discuss three categories of studies that are most closely related to us.

\textit{Route planning:} Several studies use GPS trajectories for route planning through constructing a complete route \cite{chen2010searching, chen2011discovering, yuan2010t}. Chen et al. search the $k$ Best-Connected Trajectories from a database \cite{chen2010searching} and discover the most popular route between two locations \cite{chen2011discovering}. Yuan et al. find the practically fastest route to a destination at a given departure time using historical taxi trajectories \cite{yuan2010t}.

\textit{Long-range prediction:}  Long-range prediction is studied in \cite{krumm2008markov, froehlich2008route}, where they try to predict the whole future trajectory of a moving object. Krumm proposes a Simple Markov Model that uses previously traversed road segments to predict routes in the near future \cite{krumm2008markov}. Froehlich and Krumm use previous GPS traces to make a long-range prediction of a vehicle's trajectory \cite{froehlich2008route}.

\textit{Short-range prediction:} Short-range prediction has been widely investigated \cite{xue2009traffic, jeung2008hybrid, monreale2009wherenext, morzy2007mining}, which is concerned with the prediction of only the next location. Some of these methods make prediction with only the individual movements \cite{xue2009traffic,jeung2008hybrid}, while others use the historical movements of all the moving objects \cite{monreale2009wherenext,morzy2007mining}. Xue et al. construct a Probabilistic Suffix Tree (PST) for each road using the taxi traces and propose a method based on Variable-order Markov Models (VMMs) for short-term route prediction \cite{xue2009traffic}. Jeung et al. present a hybrid prediction model to predict the future locations of moving objects, which combine predefined motion functions using the object's recent movements with the movement patterns of the object \cite{jeung2008hybrid}. Monreale et al. use the previous movements of all moving objects to build a T-pattern tree to make future location prediction \cite{monreale2009wherenext}. Morzy uses a modified version of the PrefixSpan algorithm to discover frequent trajectories and movement rules with all the moving objects' locations \cite{morzy2007mining}.

In addition to the three aforementioned categories of work, there has also appeared work on using social-media data for trajectory mining\cite{kurashima2010travel, yin2011diversified, ye2013nextmove}. 
Kurashima et al. recommend travel routes based on a large set of geo-tagged and time-stamped photographs \cite{kurashima2010travel}. Yin et al. investigate the problem of trajectory pattern ranking and diversification based on geo-tagged social media \cite{yin2011diversified}. Ye et al. utilize a mixed Hidden Markov Model to predict the category of a user's next activity and then predict a location given the category \cite{ye2013nextmove}.

\section{Preliminaries}
In this section, we will explain a few terms that are required for the subsequent discussion, and define the problem addressed in this paper.
\begin{myDef}[Sampling Location] For a given moving object $o$, it passes through a set of {\em sampling locations}, where each  sampling location refers to a point or a region (in a two-dimensional area of interest) where the position of $o$ is recorded.
\end{myDef}

For example, the positions of the cameras in the traffic surveillance system can be considered as the sampling locations.

\begin{myDef}[Trajectory Unit] For a given moving object $o$, a {\em trajectory unit}, denoted by $u$, is the basic component of its trajectory. Each trajectory unit $u$ can be represented by $\left( u.l,u.t\right)$, where $u.l$ is the id of the sampling location of the moving object at time-stamp $u.t$.
\end{myDef}

\begin{myDef}[Trajectory] For a moving object, its {\em trajectory} $T$ is defined as a time-ordered sequence of trajectory units: $<u_{1},u_{2},\ldots ,u_{n}>$.
\end{myDef}
From Definition 2, $T$ can also be represented as $<\left(u_{1}.l,u_{1}.t\right),\left(u_{2}.l,u_{2}.t\right),\ldots ,\\ \left(u_{n}.l,u_{n}.t\right)>$ where $u_{i}.t < u_{i+1}.t$ ($1\leqslant i\leqslant n-1$).

\begin{myDef}[Candidate Next Locations] For the sampling location $u_{i}.l$, we define a sampling location $u_{j}.l$ as a {\em candidate next location} of $u_{i}.l$ if a moving object can reach $u_{j}.l$ from $u_{i}.l$ directly.
\end{myDef}
The set of candidate next locations can be obtained either by prior knowledge (e.g., locations of the surveillance cameras combined with the road network graph), or by induction from historical trajectories of moving objects.



\begin{myDef} [Sampling Location Sequence] For a given trajectory $<\left(u_{1}.l,u_{1}.t\right),\\\left(u_{2}.l,u_{2}.t\right),\ldots ,\left(u_{n}.l,u_{n}.t\right)>$, its {\em sampling location sequence} refers to a sequence of sampling locations appearing in the trajectory, denoted as $<u_{1}.l,u_{2}.l,\dots, u_{n}.l>$.
\end{myDef}

\begin{myDef}[Prefix Set] For a sampling location $u_{i}.l$ and a given set of trajectories $\cal T$, its  {\em prefix set} of size $N$, denoted by  ${\cal S}_{i}^{N}$, refers to the set of sequences such that each sequence is a length $N$ subsequence that immediately precedes $u_{i+1}.l$ in the sampling location sequence of some trajectory $T\in\cal T$ .
\end{myDef}
\section{Markov Modeling}

We choose to use Markov models to solve the next location prediction problem. Specifically, a state in the Markov model corresponds to a sampling location, and state transition corresponds to moving from one sampling location to the next. 

In order to take into consideration both the collective and the individual movement patterns in making the prediction, we propose two models, a Global Markov Model (\textit{GMM}) to model the collective patterns, and a Personal Markov Model (\textit{PMM}) to model the individual patterns and solve the problem of data sparsity. They are combined using linear regression to  generate a predictor.




\subsection{Global Markov Model}


Using historical trajectories, we can train an order-$N$ \textit{GMM} to give a probabilistic prediction over the next sampling locations  for a moving object, where $N$ is a user-chosen parameter. Let $P\left(l_{i}\right)$ represents a discrete probability of a moving object arriving at sampling location $l_{i}$. The order-$N$ \textit{GMM} implies that the probability distribution $P(l^{'})$ for the next sampling location $l^{'}$ of a given moving object $o$ is independent of all but the immediately preceding $N$ locations that $o$ has arrived at:
\begin{equation}
P( l^{'}|<l_{j},\ldots,l_{i}> ) = P(l^{'}|{\cal S}_{i}^{N})
\end{equation}

For a given trajectory dataset, an order-$N$ \textit{GMM} for the sampling location $l_{i}$ can be trained in the following way.  We first construct the prefix set ${\cal S}_{i}^{N}$. Next, for every prefix in ${\cal S}_{i}^{N}$, we compute the frequency of each distinct sampling location appearing after this prefix in the dataset. These frequencies are then normalized to get a discrete probability distribution over the next sampling location.

We start with a first order \textit{GMM}, followed by a second-order \textit{GMM}, etc., until the order-$N$ \textit{GMM} has been obtained, to train a variable-order GMM. In contrast to the order-$N$ \textit{GMM}, the variable-order \textit{GMM} learns such conditional distributions with a varying $N$ and provides the means of capturing different orders of Markov dependencies based on the observed data. There exist many ways to utilize the variable-order \textit{GMM} for prediction. Here we adopt the principle of longest match. That is,  for a given sampling location sequence ending with $l_i$, we find its longest suffix match from the set of sequences in the prefix set of $l_i$.


\subsection{Personal Markov Model}


The majority of people's movements are routine (e.g., commuting), and they often have their own individual movement patterns. In addition, about 73\% of trajectories in our dataset contain only one point, but they also can reflect the characteristics of the moving objects' activities. For example, someone who lives in the east part of the city is unlikely to travel to a supermarket 50 kilo-meters away from his home. Therefore, we propose a Personal Markov Model (\textit{PMM}) for each moving object to predict next locations.

The training of PMM consists of two parts: training a variable-order Markov model for every moving object using its own trajectories of length than 1, and a zero-order Markov model for every moving object using the trajectory units.

For training the variable-order Markov model, we construct the prefix set for every moving object using its own trajectories, and then we compute the probability distribution of the next sampling locations. Specially, we iteratively train a variable-order Markov model with order $i$ ranging from 1 to $N$ using the trajectories of one moving object.


We train a zero-order Markov model using the trajectory units. 
For a moving object, let $N(l^{'})$ denotes the number of times a sampling location $l^{'}$ appears in the training trajectories. Let $L_{l^{'}}$ be the set of distinct sampling locations appearing in the training trajectories. Then we have
\begin{equation}
P(l^{'}) = \dfrac{N(l^{'})}{\sum _{l\in L_{l^{'}}}N(l)}.
\end{equation}

The zero-order Markov model  can be seamlessly integrated with the variable-order Markov model to obtain the final \textit{PMM}.

\subsection{Integration of GMM and PMM}

There are many methods to combine the results from a set of predictors. For our problem, we choose to use linear regression to integrate the two models we have proposed.

For the given $i$-th trajectory sequence, both GMM and PMM can get a vector of probabilities, $\mathbf{p}_{i}^w =\left( p_{1}^{i},p_{2}^{i},\cdots,p_{m}^{i}\right)'$ ($w=1$ for  \textit{GMM} and $w=2$ for \textit{PMM}), where $m$ is the number of the sampling locations,  and $p_{j}^{i}$ is the probability of location $j$ being the next sampling location.  We also have  a vector of indicators  $\mathbf{y}_{i}=(y_{1}^{i},y_{2}^{i},\cdots,y_{m}^{i})'$ for the $i$-th trajectory sequence, where $y_{j}^{i}=1$ if the actual next location is $j$ and 0 otherwise. We can predict $\mathbf{y}_i$ through a linear combination of the vectors generated by \textit{GMM} and \textit{PMM}:
\begin{equation}
\mathbf{\hat{y}}_i=\beta_{0} \mathbf{1}+ \sum_{w=1}^2 \beta_{w}\mathbf{p}_i^w
\end{equation}
where $\mathbf{1}$ is a unit vector, and $\beta_0$, $\beta_1$, and $\beta_2$  are the coefficients to be estimated.

Given a set of $n$ training trajectories, we can compute the optimal values of $\beta_i$ through standard linear regression that minimizes $\sum_{i=1}^n ||\mathbf{y}_i - \mathbf{\hat{y}}_i ||$, where $||\cdot||$ is the Euclidean norm. The $\beta_i$ values thus obtained can then be used for prediction. For a particular trajectory, we can predict the top $k$ next sampling locations by identifying the $k$ largest elements in the estimator $\mathbf{\hat{y}}$.

\section{Time Factor}
The movement of human beings demonstrates a great degree of temporal regularity \cite{gonzalez2008understanding,ben2010road}. In this section, we will first discuss how the movement patterns are affected by time, and then show how to improve the predictor proposed in the preceding section by taking the time factor into consideration.

\subsection{Observations and Discussions}



\begin{figure}[h]
\centering
\includegraphics[width=0.80\textwidth]{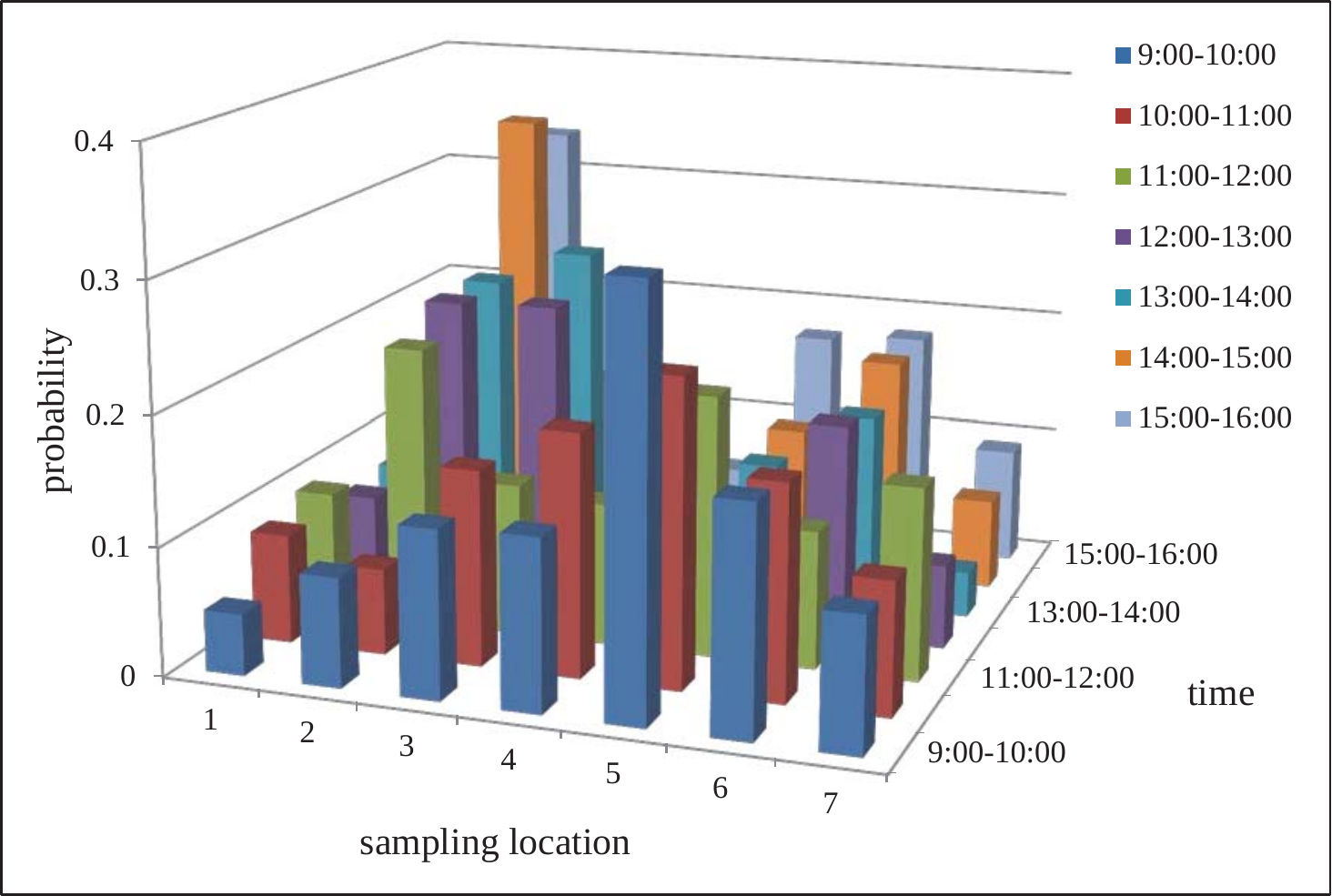}
\caption{ An example of time affecting people's movement patterns} 
\label{fig:distribution}
\end{figure}

We illustrate how time could affect people's movement patterns through Figure~\ref{fig:distribution}. In this case, for a sampling location $l$, there are seven candidate next locations, and the distributions over those locations do differ from one period to another. For instance, vehicles are most likely to arrive at the fifth location during the period from 9:00 to 10:00, whereas the most probable next location is the second for the period from 14:00 to 15:00.


Therefore, the prediction model should be made time-aware, and one way to do this is to train different models  for different time periods. In what follows, we will explore a few methods to determine the suitable time periods. Here, we choose {\em day} as the whole time span, i.e., we study how to find movement patterns within a day. However, any other units of time, such as  {\em hour}, {\em week} or {\em month}, could also be used depending on the scenario.
\subsection{Time Binning}
A straight-forward approach is to partition the time span into a given number ($M$) of equi-sized time bins, and all trajectories are mapped to those bins according to their time stamps. A trajectory spanning over more than one bin is split into smaller sub-trajectories such that the trajectory units in each sub-trajectory all fall in the same bin. We then train $M$ independent models, each for a different time bin, using the trajectories falling in each bin. Prediction is done by choosing the right model based on the time-stamp. We call this approach {\em Time Binning (TB)}.

However, this approach has some limitations: the sizes of all time bins are equal, rendering it difficult to find the correct bin sizes that fit all movement patterns in the time span, as some patterns manifest themselves over longer periods whereas others shorter. One possible improvement to {\em TB} is to start with a small bin size, and gradually merge the time bins whose distributions are considered similar by some metric. For example, in Figure~\ref{fig:distribution}, the distribution for the period from 11:00 to 12:00 is different from the one from 10:00 to 11:00; rather, it is similar to the one from 14:00 to 15:00 (e.g., they both have the maximal probability at the second sampling location).


\subsection{Distributions Clustering}

We propose a method called Distributions Clustering (\textit{DC}) to perform clustering of the time bins based on the similarities of the probability distributions in each bin. Here, the probability distribution refers to the transition probability from one location to another. Compared with \textit{TB}, the trajectories having similar probability distributions are expected to be put in one cluster, leading to clearer revelation of the moving patterns. Here, we use cosine similarity to measure the similarities between the distributions, but the same methodology still applies when other distance metrics such as the Kullback-Leibler divergence  \cite{ertoz2002new} are used.

For an object $o$ appearing at a given sampling location $l$ with a time point falling into the $i$th time bin, let  $p_{i}^{m}$ be an $m$-dimensional vector that represents the probabilities of $o$ moving from $l$ to another location, where $m$ is the total number of sampling locations. We measure the similarity of two time bins $i$ and $j$ (with respect to $o$) using the cosine similarity,
$\cos_{ij} = \left(p_{i}^{m}.p_{j}^{m}\right)/\left(\left| p_{i}^{m} \right| . \left| p_{j}^{m} \right|\right)$.
With the similarity metric defined, we can perform clustering for each sampling location $l$ on the time bins. The algorithm is detailed in Algorithm 1. The results will be a set of clusters, each containing a set of time bins, for the sampling location $l$.  

\begin{algorithm}[h]
\label{alg:dc}
\caption{DC: Detecting Q clusters for the M time bins}
\begin{algorithmic}[1]
\REQUIRE cluster number $Q$, time bins number $M$ and the probability distributions of trajectories in each time bin;
\ENSURE the clusters;
\STATE random select $Q$ time bins as the initial cluster centres;
\REPEAT
\STATE calculate the similarity of the probability distributions of trajectories in each time bin and the cluster centres;
\STATE assign each time bin to the cluster centre with the maximum similarity;
\STATE recalculate the probability distributions of trajectories in the cluster centres;
\UNTIL{clusters do not change or the maximum number of iterations has been reached}
\STATE return the clusters;
\end{algorithmic}
\end{algorithm}

For a given location $l_{i}$, we can get $Q$ clusters, defined as $C_{i}^{k}, k=1,2,\cdots,Q$. Combined with the order-$N$ Markov model, the probability distribution $P(l^{'})$ for the next sampling location $l^{'}$ of a given moving object $o$ can be computed with the formula:
\begin{equation}
P( l^{'}|<\left(u_{j}.l,u_{j}.t\right), \ldots, \left(u_{i}.l,u_{i}.t\right)> ) = P(l^{'}|C_{i}^{k},{\cal S}_{i}^{N})
\end{equation}
We then train $Q$ models with the trajectories in each cluster to form a new model \textit{NLPMM-DC} (which stands for \textit{NLPMM with Distributions Clustering}). In the new model, the sequence of just-passed locations and the time factor are both utilized by combing distributions clustering and Markov model.

\section{Performance evaluation}
We have conducted extensive experiments to evaluate the performance of the proposed \textit{NLPMM} using a real vehicle passage dataset. In this section, we will first describe the dataset and experimental settings, followed by the evaluation metrics to measure the performance. We then show the experimental results.

\subsection{Datasets and Settings }
The dataset used in the experiments consists of real vehicle passage records from the traffic surveillance system of a major metropolitan area with a 6-million population. The dataset contains 10,344,058 records during a period of 31 days (from January 1, 2013 to January 31, 2013). Each record contains three attributes, the license plate number of the vehicle, the ID of the location of the surveillance camera, and the time of vehicle passing the location. 
There are about 300 camera locations on the main roads. The average distance between a neighboring pair of camera locations is approximately 3 kilometers.


\subsection{Pre-processing}
We pre-process the dataset to form trajectories, resulting in a total of 6,521,841 trajectories. According to statistics, the trajectories containing only one point account for about 73\% of all trajectories, which testifies to the sparsity of data sampling. We choose a total of 1,760,897 trajectories with the length greater than one to calculate the number of candidate next locations for every sampling location. Due to the sparsity of camera locations, about 86.3\% of the sampling locations have more than 10 candidate next sampling locations, and the average number of candidate next locations is about 43. We predict top-$k$ next sampling locations in the experiments.

\subsection{Evaluation Metrics }
Our evaluation uses the following metrics that are widely employed in multi-label classification studies \cite{ye2011semantic}.

\newcommand{\ct}{{\cal T}}
{\textit{Prediction Coverage:}} It is defined as the percentage of trajectories for which the next location can be predicted based on the model. Let $c\left( l\right)$ be 1 if it can be predicted and 0 otherwise. Then $PreCov_{\ct}=\sum_{l\in \ct} c\left( l\right)/|\ct|$, where $|\ct|$ denotes the total number of trajectories in the testing dataset.

{\textit{Accuracy:}} It is defined as the frequency of the true next location occurring in the list of predicted next locations. Let $p\left( l\right)$ be 1 it does and 0 otherwise. Then $accuracy_{\ct}=\sum_{l\in \ct} p\left( l\right)/|\ct|$.

 {\textit{One-error:}} It is defined as the frequency of the top-1 predicted next location not being the same as the true next location. Let $e\left( l\right)$ be 0 if the top-1 predicted sampling location is the same as the true next location and 1 otherwise. Then $one-error_{\ct}=|\sum_{l\in \ct} e\left( l\right)/\ct|$.

 {\textit{Average Precision:}} Given a list of top-$k$ predicted next locations, the average precision is defined as $AvePrec_{\ct} =\sum_{l\in \ct} \left( p\left( i\right)/i\right)/|\ct|$, where $i$ denotes the position in the predicted list, and $p\left( i\right)$ takes the value of 1 if the predicted location at the $i$-th position in the list is the actual next location.
\subsection{Evaluation of NLPMM }
\begin{figure}[h]
\centering
\subfigure[]{
\includegraphics[width=0.42\textwidth,height=3.2cm]{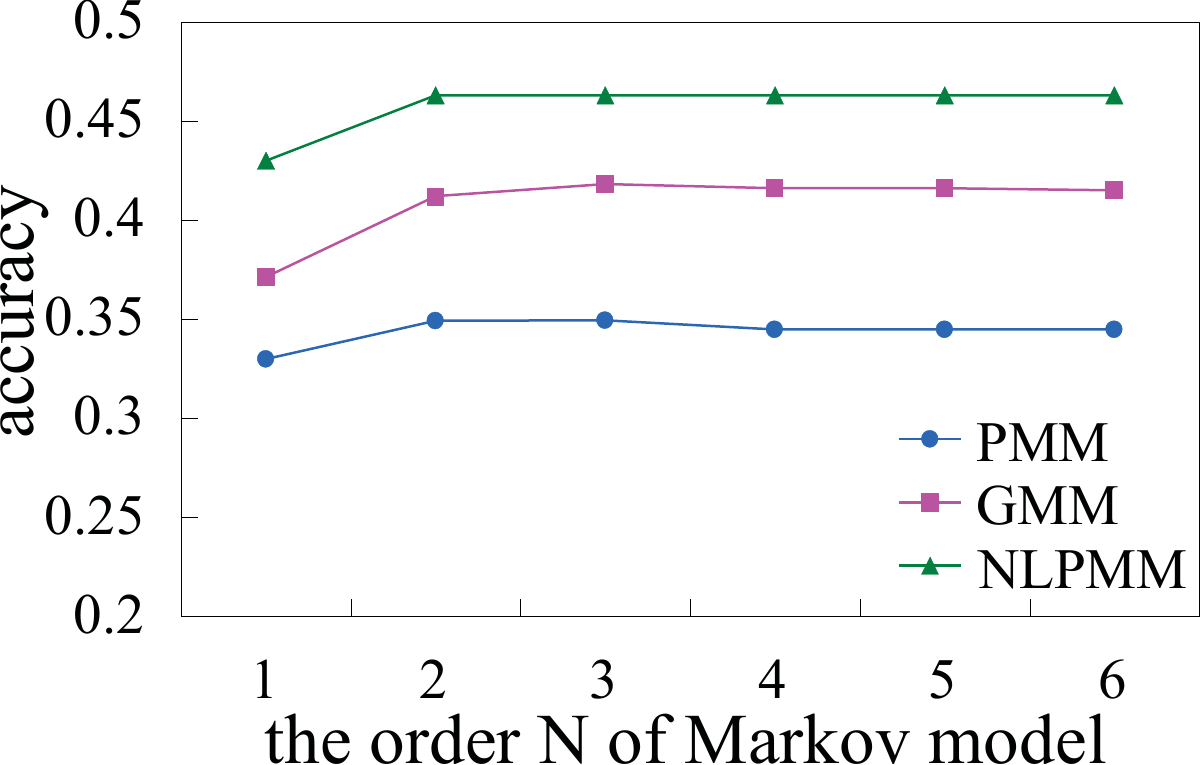}
}
\qquad
\subfigure[]{
\includegraphics[width=0.42\textwidth,height=3.2cm]{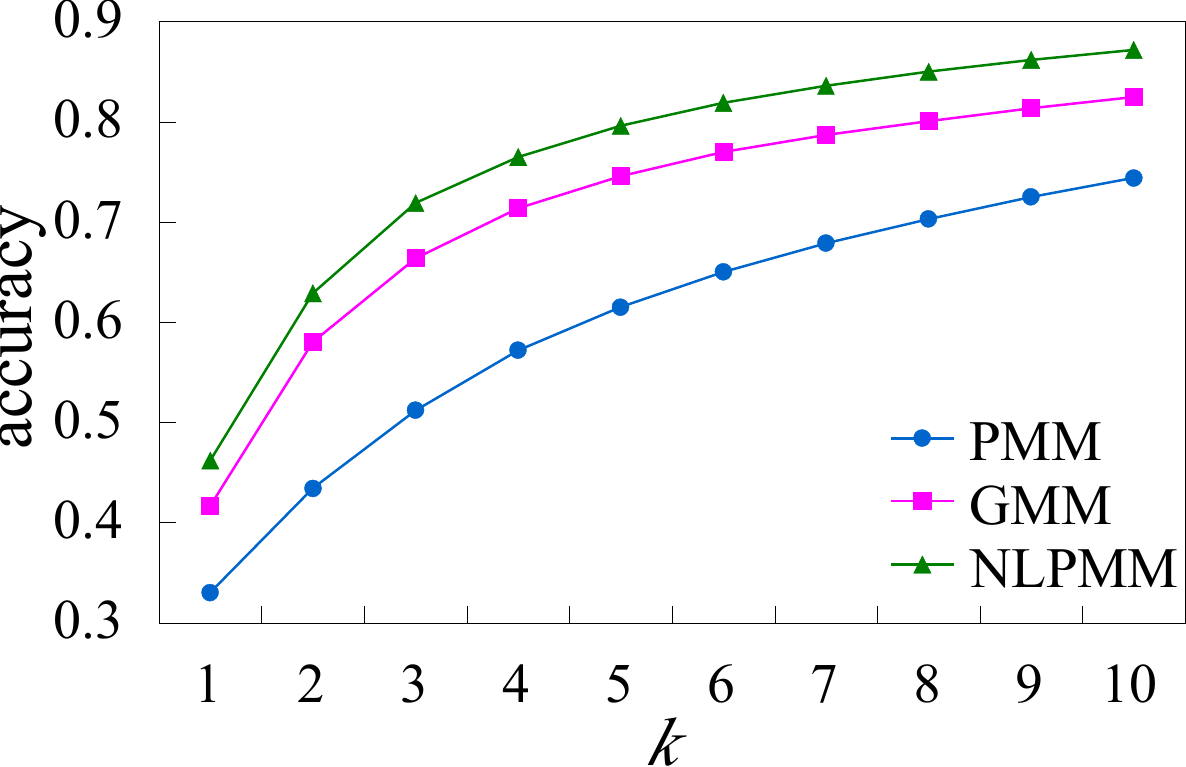}
}
\caption{Performance of PMM, GMM, and NLPMM}
\label{fig:nlpmm}
\end{figure}

We evaluate the performance of \textit{NLPMM} and its components, \textit{PMM}, and \textit{GMM}. For each experiment, we perform 50 runs and report the average of the results. First, we study the effect of the order of the Markov model by varying $N$ from 1 to 6. Figure~\ref{fig:nlpmm}(a) shows that the accuracy has an apparent improvement when the order $N$ increases from 1 to 2 for all models. The accuracy reaches the maximum when $N$ is set to 3 and remains stable as $N$ increases further. Therefore, we set $N$ to 3 in the following experiments.
Next, we evaluate the effect of top $k$ on \textit{PMM}, \textit{GMM}, and \textit{NLPMM}.  From Figure~\ref{fig:nlpmm}(b), we can observe that the accuracy of all three models improves as $k$ increases. Furthermore, the accuracy of GMM and NLPMM is significantly better than that of \textit{PMM}, and the best results are given by \textit{NLPMM}. Since the average number of candidate next locations is 43 (meaning there are 43 possibilities), the accuracy of 0.88 is surprisingly good when $k$ is set to 10. 

\subsection{Effect of the Time Factor}

We evaluate the proposed methods that take into consideration of the time factor. 
\begin{figure}[h]
\centering
\subfigure[Effect of the size of time bin]{
\includegraphics[width=0.45\textwidth,height=3.23cm]{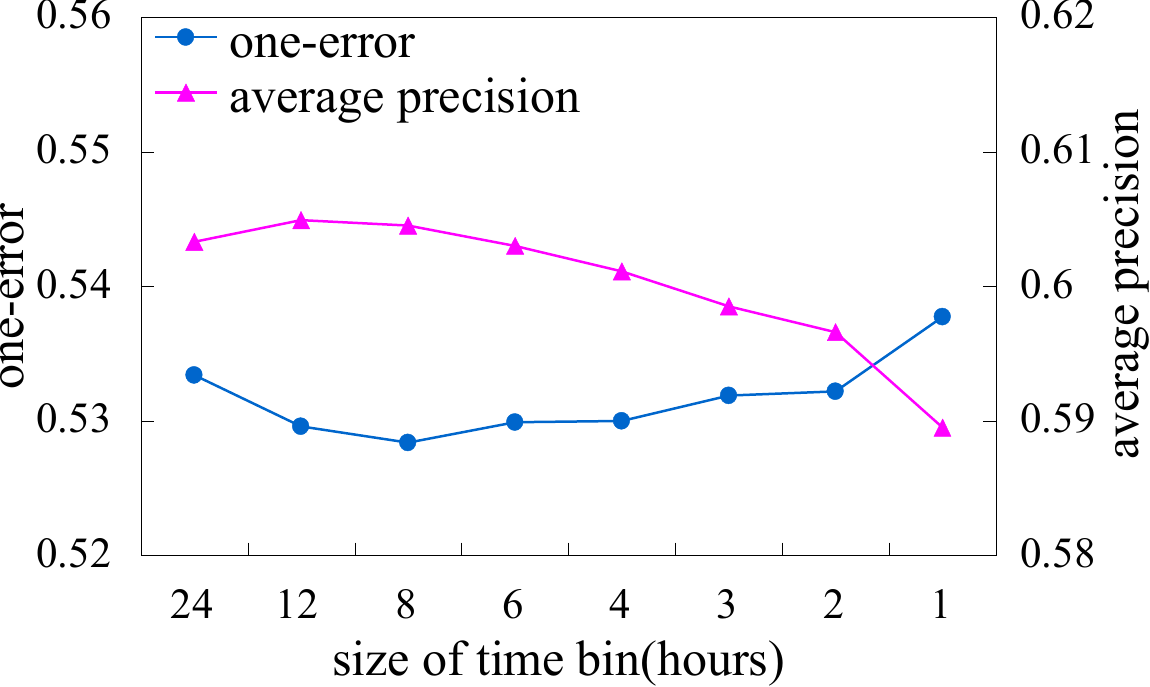}}
\quad
\subfigure[Effect of the number of clusters]{
\includegraphics[width=0.45\textwidth,height=3.23cm]{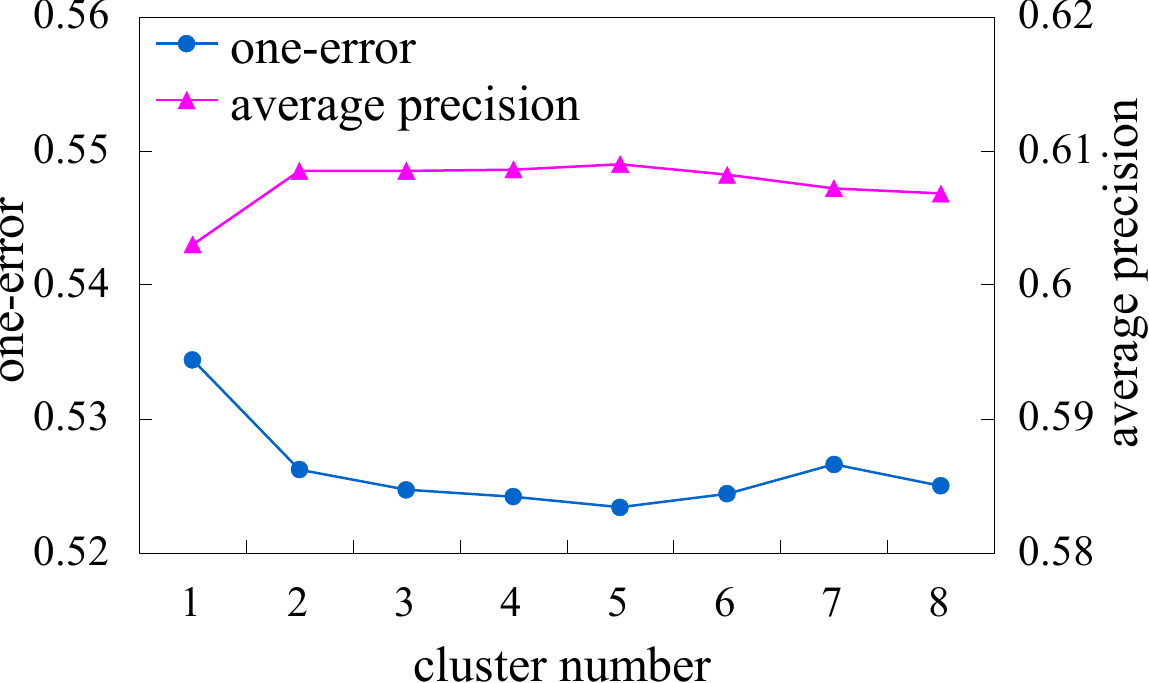}}
\caption{Effect of parameters}
\label{fig:para}
\end{figure}
Figure~\ref{fig:para}(a) shows the effect of bin size on \textit{NLPMM-TB} (which stands for \textit{NLPMM with Time Binning}). The performance of \textit{NLPMM-TB} starts to deteriorate when the bin size becomes less than 8, because when the bins get smaller, the trajectories in them become too sparse to generate a meaningful collective pattern.  Figure~\ref{fig:para}(b) shows the effect of the number of clusters on \textit{NLPMM-DC} (which stands for \textit{NLPMM with Distributions Clustering}).  When it is set to 1, the model is the same as \textit{NLPMM}. The one-error rate declines and the average precision improves as the number increases from 1 to 5. When it continues to increase, the result starts to get worse. This is because having too many or too few clusters with either hurt the cohesiveness or the separation of the clusters. 

We evaluate the performance of \textit{NLPMM}, \textit{NLPMM-TB} and \textit{NLPMM-DC} using one-error and average precision. The results are shown in Table~\ref{tab:time}. \textit{NLPMM-TB} and \textit{NLPMM-DC} perform better than \textit{NLPMM}, which is because we can get a more refined model by adding the time factor and generate more accurate predictions. \textit{NLPMM-DC} performs best, validating the effectiveness of the method of distributions clustering. It will be used in the following comparison with alternative methods.
\begin{table}
\caption{one-error and average precision of different models}
\centering
\begin{tabular}{|p{0.25\textwidth}<{\centering}|p{0.2\textwidth}<{\centering}|p{0.2\textwidth}<{\centering}|p{0.2\textwidth}<{\centering}|}
\hline
 & NLPMM & NLPMM-TB & NLPMM-DC \\
\hline
one-error & 53.8\% & 53.0\% & \textbf{52.3\%}  \\
\hline
average precision & 60.2\% & 60.5\% & \textbf{60.9\%}  \\
\hline
\end{tabular}
\label{tab:time}
\end{table}	
\subsection{Comparison with existing methods}

We compare the proposed \textit{NLPMM-DC} with the start-of-the-art approaches \textit{VMM} \cite{xue2009traffic} and \textit{WhereNext} \cite{monreale2009wherenext}. \textit{VMM} uses individual trajectories to predict the next locations, whereas \textit{WhereNext} uses all available trajectories to discover collective patterns. In this experiment, we predict top-$1$ next sampling location. The parameters of \textit{VMM} are set as follows: memory length $N$=3, $\sigma$=0.3, and $N_{\min }$=1. For \textit{WhereNext}, the support for constructing T-pattern tree is set as 20. For the \textit{NLPMM-DC}, the setting is that the order $N$ = 3 and the number of clusters is set at 5.

\begin{figure}
\centering
\subfigure[]{
\includegraphics[width=0.42\textwidth,height=3.2cm]{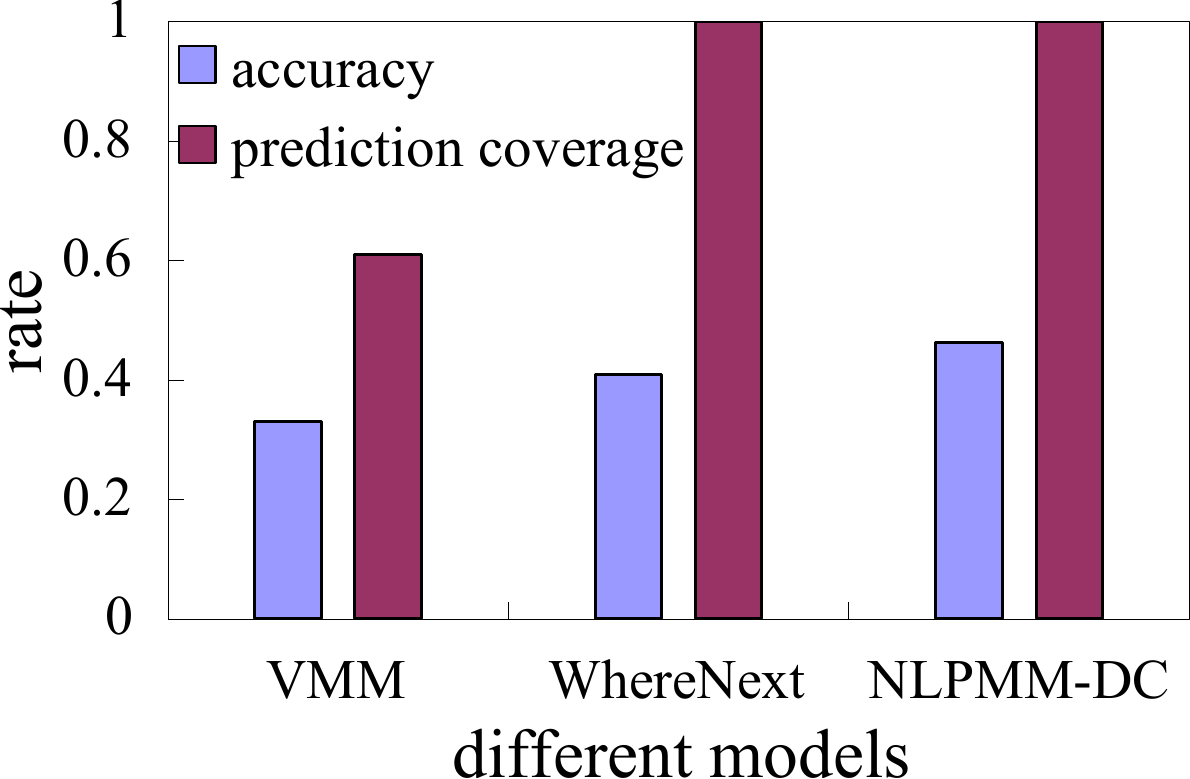}}
\qquad
\subfigure[]{
\includegraphics[width=0.42\textwidth,height=3.2cm]{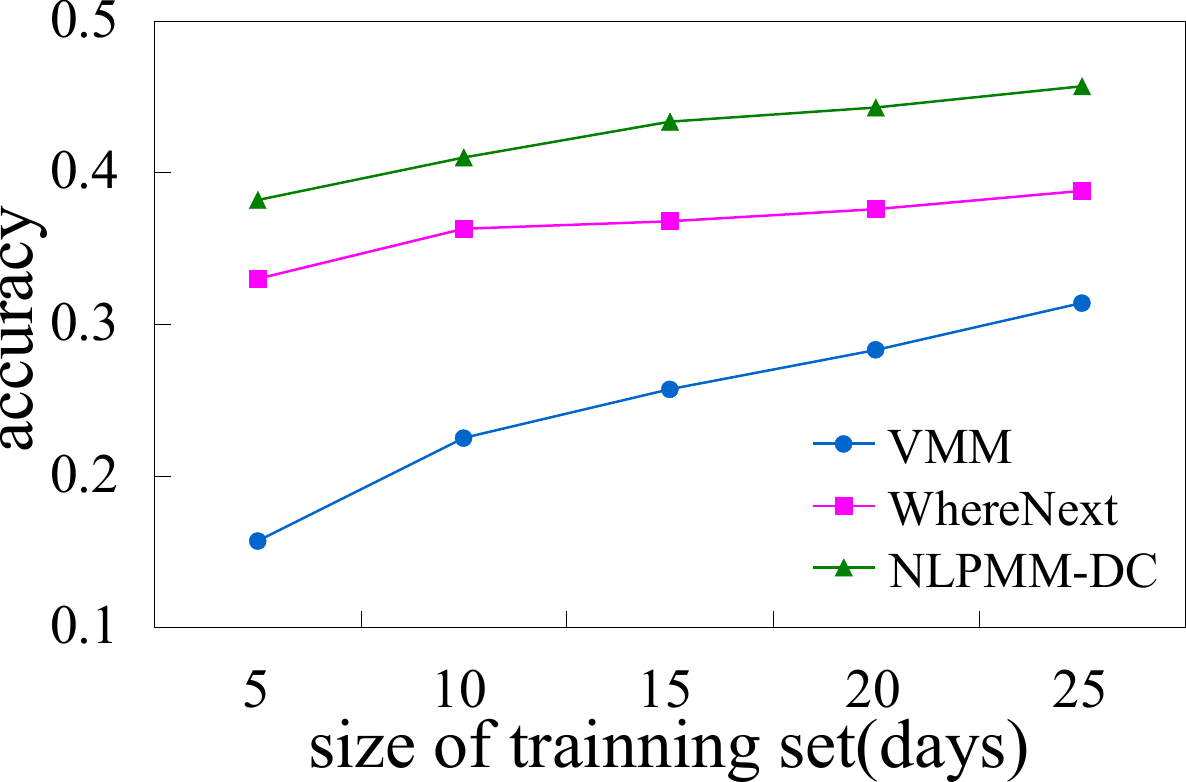}}
\caption{Performance comparison of \textit{NLPMM-DC}, \textit{VMM}, and \textit{WhereNext}}
\label{fig:baseline}
\end{figure}
Figure~\ref{fig:baseline} shows the performance comparison of \textit{NLPMM-DC}, \textit{VMM} and \textit{WhereNext} in terms of prediction coverage and accuracy. As shown in Figure~\ref{fig:baseline}(a), \textit{NLPMM-DC} performs the best, which can be attributed to the combination of individual and collective patterns as well as the consideration of time factor. Figure~\ref{fig:baseline}(b) shows that the accuracy of each model improves as the size of training set increases. It is worth mentioning that \textit{NLPMM-DC} performs better than \textit{VMM} and \textit{WhereNext} in terms of accuracy for any training set size.
\section{Conclusions}
In this paper, we have proposed a Next Location Predictor with Markov Modeling to predict the next sampling location that a moving object will arrive at with a given trajectory sequence. The proposed \textit{NLPMM} consists of two models: Global Markov Model and Personal Markov Model. Time factor is also added to the models and we propose two methods to partition the whole time span into periods of finer granularities, including Time Binning and Distributions Clustering. New time-aware models are trained accordingly. We have evaluated the proposed models using a real vehicle passage record dataset. The experiments show that our predictor significantly outperforms the state-of-the-art methods (\textit{VMM} and \textit{WhereNext}).

\bibliographystyle{splncs03}
\bibliography{wenxian}

\end{document}